\documentclass[pdflatex,sn-mathphys-num]{sn-jnl}

\usepackage{graphicx}%
\usepackage{multirow}%
\usepackage{amsmath,amssymb,amsfonts}%
\usepackage{amsthm}%
\usepackage{mathrsfs}%
\usepackage[title]{appendix}%
\usepackage{xcolor}%
\usepackage{textcomp}%
\usepackage{manyfoot}%
\usepackage{booktabs}%
\usepackage{algorithm}%
\usepackage{algorithmicx}%
\usepackage{algpseudocode}%
\usepackage{listings}%

\usepackage{subcaption}%
\usepackage{tabularx} %
\usepackage{tikz} 
\usepackage[left]{lineno}

\theoremstyle{thmstyleone}%
\theoremstyle{thmstyletwo}%

\theoremstyle{thmstylethree}%

\begin{document}
\title[Article Title]{Physics-informed Partitioned Coupled Neural Operator for Complex Networks }


\author[1]{\fnm{Weidong} \sur{Wu}}\email{wdwu229@163.com}

\author*[1]{\fnm{Yong} \sur{Zhang}}\email{yongzh401@126.com}

\author[1]{\fnm{Lili} \sur{Hao}}\email{haolihaoli@cumt.edu.cn}

\author[2]{\fnm{Yang} \sur{Chen}}\email{fedora.cy@gmail.com}
\author[3]{\fnm{Xiaoyan} \sur{Sun}}\email{xysun78@126.com}
\author[4]{\fnm{Dunwei} \sur{Gong}}\email{dwgong@vip.163.com}

\affil*[1]{\orgdiv{School of Information and Control Engineering}, \orgname{China University of Mining and Technology}, \orgaddress{\street{No. 1, Daxue Road}, \city{Xuzhou}, \postcode{221116}, \state{Jiangsu}, \country{China}}}

\affil[2]{\orgdiv{School of Electrical Engineering}, \orgname{China University of Mining and Technology}, \orgaddress{\street{No. 1, Daxue Road}, \city{Xuzhou}, \postcode{221116}, \state{Jiangsu}, \country{China}}}

\affil[3]{\orgdiv{School of Artificial Intelligence and Computer Science}, \orgname{Jiangnan University}, \orgaddress{\street{No. 1800, Lihu Avenue}, \city{Wuxi}, \postcode{214122}, \state{Jiangsu}, \country{China}}}

\affil[4]{\orgdiv{School of Information Science and Technology}, \orgname{Qingdao University of Science and technology}, \orgaddress{\street{NO. 99, Songling Road}, \city{Qingdao}, \postcode{266061}, \state{Shandong}, \country{China}}}


\abstract{Physics-Informed Neural Operators provide efficient, high-fidelity simulations for systems governed by partial differential equations (PDEs). However, most existing studies focus only on multi-scale, multi-physics systems within a single spatial region, neglecting the case with multiple interconnected sub-regions, such as gas and thermal systems. To address this, this paper proposes a Physics-Informed Partitioned Coupled Neural Operator (PCNO) to enhance the simulation performance of such networks. Compared to the existing Fourier Neural Operator (FNO), this method designs a joint convolution operator within the Fourier layer, enabling global integration capturing all sub-regions. Additionally, grid alignment layers are introduced outside the Fourier layer to help the joint convolution operator accurately learn the coupling relationship between sub-regions in the frequency domain. Experiments on gas networks demonstrate that the proposed operator not only accurately simulates complex systems but also shows good generalization and low model complexity.}

\keywords{  Physics-informed Neural Operator, Complex Networks, Partial Differential Equations, Natural Gas Pipeline Networks}



\maketitle

\section{Introduction}\label{sec1}
In recent years, as scientific discoveries and engineering design problems have become increasingly complex, traditional experimental methods have struggled to effectively address these challenges due to limitations in time and cost. The rapid development of artificial intelligence technologies introduces new approaches for modelling and predicting complex systems  \cite{bib1,bib2}. By replacing costly and time-consuming physical experiments with computer simulations, scientists can more efficiently infer complex phenomena, thereby accelerating scientific simulations and engineering designs. In fact, many physical phenomena occurring in continuous domains, such as physicochemical reactions \cite{bib4,bib6}, changes in the microstructure of materials \cite{bib3,bib5,bib7}, and fluid flow \cite{bib8}, can be described by governing laws like partial differential equations (PDEs). However, traditional numerical methods, such as finite difference \cite{bib9} finite element \cite{bib10}, finite volume \cite{bib11}, and spectral methods \cite{bib12}, often struggle to efficiently achieve high-fidelity simulations for complex systems governed by known laws, as they are constrained by factors such as grid resolution, model complexity, and computational resources. Moreover, for physical problems requiring finer scales, these methods are unable to directly leverage observational data and often demand substantial computational resources \cite{bib13}. Therefore, there is an urgent need to develop more advanced simulation tools or methods to enhance the accuracy and efficiency of simulations for practical problems and meet the demands of complex scientific and engineering challenges.

As a novel deep learning framework, neural operators \cite{bib14,bib28} have been proven to be an highly efficient alternative for solving partial differential equations. Unlike traditional numerical methods, neural operators can learn the continuous mappings of a system from discrete data, offering accurate predictions with consistent generalization across different discretization conditions. Now several efficient neural operators have been proposed, such as the Deep Operator Network (DeepONet) \cite{bib16}, Graph Kernel Network \cite{bib21}, Fourier Neural Operator (FNO) \cite{bib17}, and Laplace Neural Operator (LNO) \cite{bib18}. These operators have demonstrated excellent performance in abstract PDE systems and complex scenarios, including irregular grids \cite{bib22,bib23}, multi-scale systems \cite{bib26,bib29}, and transfer learning \cite{bib24,bib25}. Neural operators have also been applied to a wide range of practical problems, such as battery degradation modelling \cite{bib31}, computer vision tasks \cite{bib32}, dynamical manifold reconstruction \cite{bib33}, and multiphase flow simulations \cite{bib34}. However, the aforementioned achievements are primarily focused on systems where physical phenomena occur within a single spatial domain. Research on complex networked systems involving multiple coupled sub-regions remains relatively limited.

In many complex systems, multiple sub-regions may coexist, with each sub-region governed by partial differential equations characterized by different parameters. These equations are coupled through interface conditions, forming what is known as a "partitioned coupled system." For example, in natural gas and thermal networks, the physical characteristics of pipelines (such as length and diameter) vary, and quantities like fluid velocity, pressure, and temperature are governed by different equations in each segment. These quantities are coupled and exchanged at the pipeline connection points, ensuring continuity and balance across the network. Other typical examples include urban traffic networks, river networks, and urban wastewater treatment networks. Although scholars have conducted research on these complex systems and proposed neural operators for vehicle density prediction \cite{bib35}, geometric-adaptive flood forecasting systems \cite{bib36}, and urban water clarification hydrodynamics and
particulate matter transport \cite{bib37}, these methods are still primarily focused on simplified single-region scenarios and have not fully addressed the complexity of partitioned coupled systems. Existing neural operators face significant challenges when dealing with multiple coupled sub-regions.

To address the limitations of existing neural operators in handling multi-domain complex network systems, this paper proposes a physics-informed partitioned coupled neural operator (PCNO). First, a joint convolution operator is introduced within the framework of the Fourier neural operator to learn the coupling relationships among multiple sub-regions, enabling global integration across the regions. Second, considering the varying sizes of the sub-regions and the non-periodicity of the input functions, grid alignment layers are proposed to ensure that the joint convolution operator accurately computes global integration in the frequency domain. Additionally, to improve model accuracy and mitigate the optimization challenges faced by neural operators when relying solely on physical constraints, a training mechanism that incorporates both physics knowledge and low-resolution data is employed to help the neural operator more efficiently identify the optimal model parameters. Finally, the proposed neural operator is applied to a natural gas pipeline network, where its effectiveness is validated through experiments, and its characteristics are analyzed.

\section{Results}\label{sec2}
The experiments in this paper analyse the accuracy and generalization capability of the proposed PCNO (see Fig.~\ref{fig2}) using a natural gas network system consisting of three pipelines (see Fig.~\ref{fig1}). Furthermore, we investigate the factors influencing the accurate simulation of complex networks by evaluating the effectiveness of the grid alignment layer and joint convolution operator under various conditions. The specific experimental setup is provided in Section~\ref{sec2.1} of the Methods, while the application of PCNO to the natural gas network is detailed in the Supplementary Section~\ref{sup2}.

\subsection{Generalization Performance of PCNO}\label{sec2.2}

This experiment evaluates the generalization performance of the PCNO trained with low-resolution data and/or PDEs on high-resolution datasets. The training process utilizes 500 PDE instances and 500 low-resolution simulation datasets, with discretization steps of 320s × 200m and 640s × 400m, respectively. To further assess the model's performance, tests are conducted on datasets with 200 samples across different resolutions: 640s × 400m, 320s × 200m, and 160s × 100m. The performance is evaluated using the relative L2 error to measure the operator's effectiveness across different resolutions.

\begin{figure}[h]
\centering
    \begin{subfigure}[b]{\textwidth}
        \centering
        \includegraphics[width=0.7\textwidth]{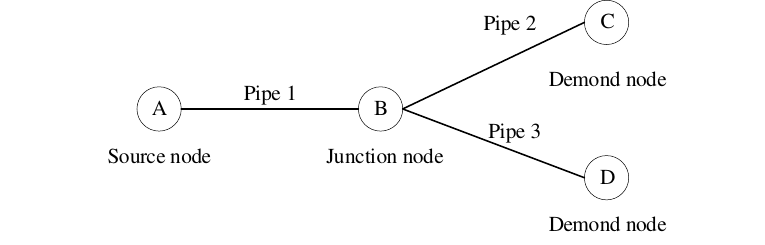} 
        \label{fig1a}
    \end{subfigure}

    \begin{subfigure}[b]{\textwidth}
        \centering
        \includegraphics[width=0.9\textwidth]{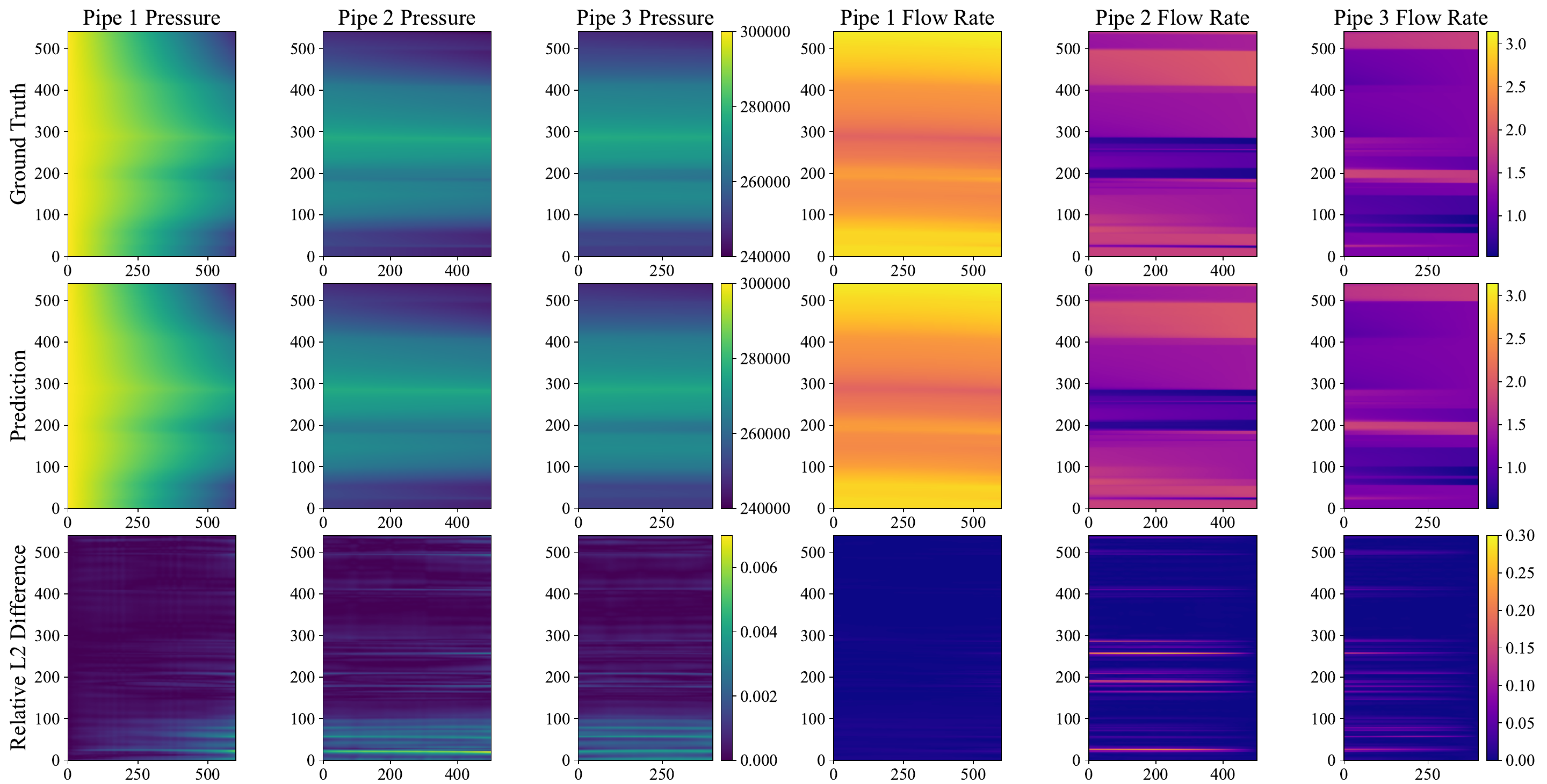} 
        \label{fig1b}
    \end{subfigure}
\caption{Spatiotemporal distribution of flow rate and pressure in the natural gas network at a resolution of 160s*100m. The x-axis represents the pipeline positions, the y-axis denotes time, and the colour indicates the magnitude of flow or pressure. The top part of the figure shows the connection diagram of the pipelines.}
\label{fig1}
\end{figure}

Figure~\ref{fig1} presents the spatiotemporal distribution of flow and pressure in the gas network under the previously unseen resolution of 160s × 100m. Pipeline 1 is connected to the source node, with its left boundary pressure fixed at 0.3 MPa. Pipelines 2 and 3 are connected to demand nodes, with their right boundaries governed by flow rates, where sudden changes in flow are observed. The states of the remaining pipeline segments are solved using the PCNO. From the error subplots of pipelines 2 and 3 in Fig.~\ref{fig1}, it can be observed that the overall error remains within an acceptable range, while the error is slightly larger during flow discontinuities compared to other moments. Additionally, the pressure changes slowly, with the error consistently maintained at a low level.

\begin{table}[h]
  \caption{Relative L2 Error of the PCNO at Different Resolutions. The operator is trained using different amounts of observational data (640s × 400m) and/or PDE data (320s × 200m). The test set resolutions are 640s × 400m, 320s × 200m, and 160s × 100m, where the time step is measured in seconds and the spatial step in meters.}
  \centering
  \setlength{\tabcolsep}{2pt}
    \begin{tabular}{cc|cc|cc|cc}
    \toprule
    \multicolumn{2}{c|}{Training} & \multicolumn{2}{c|}{640s*400m} & \multicolumn{2}{c|}{320s*200m} & \multicolumn{2}{c}{160s*100m} \\
    \midrule
    \multicolumn{1}{c}{Data} & \multicolumn{1}{c|}{PDE} & Flow  & Pressure & Flow  & Pressure & Flow  & Pressure \\
    \midrule
    500   & 0     & \textbf{0.0028+0.0045} & 0.0003+0.0006 & \textbf{0.0065+0.0148} & \textbf{0.0008+0.0027} & 0.0078+0.0200 & \textbf{0.0010+0.0036} \\
    500   & 500   & 0.0031+0.0051 & \textbf{0.0003+0.0005} & 0.0065+0.0150 & \textbf{0.0008+0.0027} & \textbf{0.0075+0.0201} & \textbf{0.0010+0.0036} \\
    375   & 500   & 0.0037+0.0056 & 0.0005+0.0007 & 0.0069+0.0154 & 0.0009+0.0028 & 0.0079+0.0204 & 0.0011+0.0037 \\
    250   & 500   & 0.0051+0.0080 & 0.0009+0.0021 & 0.0081+0.0172 & 0.0013+0.0037 & 0.0090+0.0219 & 0.0015+0.0044 \\
    125   & 500   & 0.0102+0.0171 & 0.0027+0.0061 & 0.0127+0.0235 & 0.0030+0.0071 & 0.0134+0.0268 & 0.0031+0.0073 \\
    0     & 500   & 0.0333+0.0487 & 0.0098+0.0161 & 0.0347+0.0519 & 0.0098+0.0162 & 0.0352+0.0541 & 0.0099+0.0162 \\
    \bottomrule
    \end{tabular}%
  \label{tab1}%
\end{table}%

Based on 200 test datasets, Table~\ref{tab1} provides the relative L2 error results of the PCNO trained with different combinations of observational data and PDE instances across varying resolutions. The values in the table represent the point-wise mean and standard deviation of the error. The results show that neural operators trained with either observational data alone or the combination of observational data and PDE instances exhibit smaller errors than those trained with fewer data. Further analysis reveals that, although the operator trained solely with data performs better in flow prediction at the seen resolution (640s × 400m), the operator trained with both physical information and observational data achieves superior performance at the unseen resolution (160s × 100m). This indicates that the inclusion of physical information is essential for high-resolution predictions, though it offers limited improvement to the PCNO’s overall performance.

Furthermore, Table~\ref{tab1} also shows that as the proportion of observational data in the training set decreases, the model's error increases. When the PCNO is trained exclusively with PDE instances, it has the highest error. This indicates that the PDE constraints in the network system are too complex to be effectively handled by soft constraints alone, thus limiting the effectiveness of the operator’s training.

\subsection{Effectiveness of the Grid Alignment Layer}\label{sec2.3}
According to the analysis in \textbf{Section 3.3}, the grid alignment layer can align discretized grids of varying sizes from different regions, assisting the joint convolution operator in performing unified convolution under the same set of Fourier basis functions, and enhancing the operator’s adaptability to non-periodic boundary conditions. In this experiment, 1,000 data samples and PDE instances are used to train PCNOs with different grid alignment parameters. The performance is evaluated using the relative L2 error on 200 test data at a resolution of 640s × 400m. The fill ratios for the temporal dimension ${r_t}$ are set to 0, 0.05, and 0.1, while the spatial fill ratios ${r_x}$ are set to 0, 0.001, 0.05, and 0.1. It is worth noting that when ${r_x} = 0.001$, those smaller grids in the spatial dimension are filled until their spatial length matches that of the largest grid in the spatial dimension.

\begin{table}[h]
  \centering
  \caption{The relative L2 error of PCNO for grid alignment layers with different parameters. ${Z_{t,\max }}$ and ${Z_{x,\max }}$ represent the cutoff frequencies of the Fourier layers in the temporal and spatial dimensions, respectively. The fill ratios for the temporal dimension ${r_t}$ are set to 0, 0.05, and 0.1, while the fill ratios for the spatial dimension ${r_x}$ are set to 0, 0.001, 0.05, and 0.1.}
  \setlength{\tabcolsep}{2pt}
    \begin{tabular}{c|cc|cc|cc}
    \toprule
    \multicolumn{7}{c}{${Z_{t,\max }} = 25$, ${Z_{x,\max }} = 25$} \\
    \midrule
    \multicolumn{1}{c|}{
            \begin{tikzpicture}[scale=0.1]
                \draw[thick] (0,) -- (1,0); 
                \node[anchor=south] at (0.5,1) {${r_t}$}; 
                \node[anchor=east] at (0,0.5) {${r_x}$};
            \end{tikzpicture}
    
    } & \multicolumn{2}{c|}{0} & \multicolumn{2}{c|}{0.05} & \multicolumn{2}{c}{0.1} \\
    \midrule
    0     & 0.0033+0.0053 & 0.0003+0.0006 & 0.0035+0.0057 & 0.0004+0.0008 & 0.0038+0.0062 & 0.0005+0.0008 \\
    0.0001 & \textbf{0.0031+0.0051} & \textbf{0.0003+0.0005} & 0.0038+0.0061 & 0.0005+0.0009 & 0.0038+0.0064 & 0.0005+0.0008 \\
    0.05  & 0.0032+0.0051 & 0.0004+0.0006 & 0.0039+0.0061 & 0.0006+0.0012 & 0.0039+0.0063 & 0.0005+0.0010 \\
    0.1   & 0.0034+0.055 & 0.0004+0.0006 & 0.0036+0.0057 & 0.0005+0.0009 & 0.0036+0.0059 & 0.0004+0.0006 \\
    \midrule
    \multicolumn{7}{c}{${Z_{t,\max }} = 45$, ${Z_{x,\max }} = 25$} \\
    \midrule
    0.0001 & 0.0018+0.0027 & 0.0003+0.0006 & 0.0018+0.0026 & 0.0003+0.0005 & \textbf{0.0018+0.0025} & \textbf{0.0002+0.0003} \\
    \midrule
    \multicolumn{7}{c}{${Z_{t,\max }} = 68$, ${Z_{x,\max }} = 25$} \\
    \midrule
    0.0001 & 0.0011+0.0015 & 0.0003+0.0003 & 0.0011+0.0015 & 0.0002+0.00004 & \textbf{0.0011+0.0014} & \textbf{0.0002+0.0003} \\
    \bottomrule
    \end{tabular}%
  \label{tab2}%
\end{table}%

As shown in Table~\ref{tab2}, when the cutoff frequencies of the Fourier layers are set to be ${Z_{t,\max }} = {Z_{x,\max }} = 25$ and ${r_t} = 0$, increasing the spatial fill ratio decreases the relative L2 error at first, but the error increases again after reaching its minimum at ${r_x} = 0.001$. This indicates that setting identical Fourier basis functions across different regions is more conducive to accurate joint convolution computation. However, since the cutoff frequency is fixed, further increasing the grid size will reduce the amount of frequency information captured from the input during joint convolution in the Fourier layer. Moreover, when the cutoff frequencies are increased to ${Z_{t,\max }} = 45$ and ${Z_{t,\max }} = 68$, the increase of the temporal fill ratio slightly reduces the error. This suggests that under appropriately chosen cutoff frequencies, grid filling can better mitigate the impact of non-periodic boundary conditions, enhancing the operator’s adaptability.

\subsection{Effectiveness of the Joint Convolution }\label{sec2.4}
According to the analysis in method Section~\ref{sec3.4}, the joint convolution operator in the Fourier layer plays a key role in integrating information across multiple regions. This includes the inter-regional convolution ${\mathcal{K}_{l1}}$ and the fusion of channel information ${\mathcal{K}_{l2}}$. To verify the effectiveness of the proposed joint convolution operator, we discussed the following four integral operators:

1) PCNO-C: ${{\mathcal K}_{l2}}$ replaced by ${\mathcal{K}_{l2}} \in {\mathbb{C}^{{d_{l + 1}}*{d_l}}}$, meaning that identical weight parameters are used for each region when fusing channel information.

2) PCNO-3D: Both ${{\mathcal K}_{l1}}$ and ${{\mathcal K}_{l2}}$ are removed and replaced by ${\mathcal K} \in {{\Bbb C}^{{d_{l + 1}}*{d_l}*{d_{{Z_{t,\max }}}}*{d_{{Z_{x,\max }}}}*{d_{{Z_{e,\max }}}}}}$. Unlike the PCNO, which applies Fourier transforms only over temporal and spatial dimensions, PCNO-3D extends the transform to include the regional dimension to assess its impact on performance.

3)	FNO-2D\cite{bib17}: This method is a standard operator without a grid alignment layer, which performs convolution within regions but cannot handle interactions between regions.

4)	FNO-3D: This is a modified version of FNO-3D\cite{bib17} with the addition of a grid alignment layer, allowing the regional dimension to participate in the Fourier transformation. However, compared with PCNO-3D, the complexity of its frequency parameters increases.

For detailed information on these operators, please refer to the supplementary baseline Section~\ref{sup3}.

\begin{table}[h]
  \centering
  \caption{Relative L2 errors of five operators on datasets with different resolutions. The rightmost column shows the model size of each operator, measured in megabytes (MB).}
  \setlength{\tabcolsep}{2pt}
    \begin{tabular}{c|cc|cc|cc|c}
    \toprule
    \multirow{2}[4]{*}{Method} & \multicolumn{2}{c|}{640s*400m} & \multicolumn{2}{c|}{320s*200m} & \multicolumn{2}{c|}{160s*100m} & \multicolumn{1}{c}{model size} \\
\cmidrule{2-8}    \multicolumn{1}{r|}{} & Flow  & Pressure & Flow  & Pressure & Flow  & Pressure & \multicolumn{1}{c}{(MB)} \\
    \midrule
    PCNO  & \textbf{0.0031+0.0051} & \textbf{0.0003+0.0005} & \textbf{0.0065+0.0150} & \textbf{0.0008+0.0027} & \textbf{0.0075+0.0201} & \textbf{0.0010+0.0036} & 3.83 \\
    PCNO-C & 0.0038+0.0061 & 0.0004+0.0007 & 0.0072+0.0160 & 0.0009+0.0029 & 0.0082+0.0210 & 0.0012+0.0037 & 2.33 \\
    PCNO-3D & 0.0063+0.0100 & 0.0015+0.0041 & 0.0092+0.0177 & 0.0019+0.0054 & 0.0103+0.0221 & 0.0021+0.0059 & 1406.8 \\
    FNO-2D  & 0.0041+0.0065 & 0.0010+0.0026 & 0.0074+0.0165 & 0.0014+0.0041 & 0.0083+0.0214 & 0.0016+0.0047 & 469 \\
    FNO3D & 0.0073+0.0116 & 0.0018+0.0046 & 0.0116+0.0195 & 0.0024+0.0058 & 0.0129+0.0236 & 0.0026+0.0063 & 2813.06 \\
    \bottomrule
    \end{tabular}%
  \label{tab3}%
\end{table}%

Table~\ref{tab3} presents the performance across different resolutions and model parameter counts of the operators. Compared to FNO-2D, which does not account for inter-regional coupling, and PCNO-3D and FNO-3D, which incorporate Fourier transforms over the regional dimension, the proposed PCNO demonstrates significant advantages at all resolutions. Additionally, compared to PCNO-C, which disregards regional differences, the proposed PCNO further improves accuracy. Moreover, PCNO reduces the model parameters by 122 times and 734 times compared to FNO-2D and FNO-3D respectively, while achieving higher accuracy.

\section{Discussion}\label{sec3}
PCNO is a physics-informed neural operator characterized by high generalization ability, accuracy, and low model complexity. We systematically analysed the performance of PCNO on a natural gas network in the previous section, in which we examined how to optimize its learning process to enhance simulation accuracy and generalization. The results indicate that the addition of physical information offers only limited improvements in both accuracy and generalization, particularly in complex network environments. This finding differs from the conclusions in \cite{bib39}. We speculate that this discrepancy arises from the fact that the PDE constraints within complex networks are far more challenging than those associated with classical physics problems, such as Darcy flow or Navier-Stokes equations. This suggests that, in systems governed by complex PDE constraints, data plays also an important role in the learning process of the operator, like physical constraints.

Additionally, we explored the conditions for accurately calculating global integrals across all sub-regions. From the comparison results with existing algorithms, we identified the factors within the joint convolution operator that influence the accuracy of complex network simulations. Our results indicate that accounting for coupling relationships between sub-regions significantly improves model’s accuracy. When learning such couplings, some key factors must be considered, including the consistency of information between regions, the methods used to integrate regional information, and the strategies for processing information within each region.

Overall, the proposed PCNO, which accounts for inter-regional coupling, demonstrates superior suitability for simulating complex networks compared to existing neural operators. The introduction of the grid alignment layer ensures the consistency of information between regions, while the proposed joint convolution operator not only performs integrals within individual regions but also effectively integrates information across regions. These operations align naturally with the characteristics of complex networks, where different regions are governed by distinct PDEs and are coupled through interfaces.

However, our model still has limitations in terms of flexibility. The parameter settings of the joint convolution operator depend on the number of sub-regions within the complex network. When the network encounters unexpected faults or topological changes that alter the number of sub-regions, the PCNO requires retraining. Additionally, the current model is not directly applicable through transfer learning to networks of the same type but with different scales. 

PCNO holds great potential for applications in various complex network systems. For example, in areas such as thermal networks, traffic flow monitoring, and flood forecasting in river networks, effective training can be achieved by combining low-resolution data from numerical solvers or field measurements with physical information.

Further research is required to fully explore the potential of PCNO. First, methods for automatic weight adjustment should be developed to address the increasingly complex PDE constraints in network systems. Second, for the cases involving irregular networks and other specialized complex systems, some more effective improvements, such as, incorporating geometric adaptivity \cite{bib56} into the proposed method, are needed to improve the model’s robustness and applicability across different scenarios. Additionally, future research should explore methods for integrating inter-regional information, ensuring that the calculation of global integrals is independent of the number of sub-regions. This will enhance the adaptability and robustness of PCNO in dynamic complex networks.

\section{Methods}\label{sec4}
\subsection{Overview}\label{sec3.0}
The proposed Partitioned Coupled Neural Operator aims to enhance the accuracy and generalization of complex network system simulations by effectively managing the coupling relationships between sub-regions. Although each sub-region within a complex network is governed by PDEs with distinct parameters, they remain closely interconnected. This study focuses on designing an operator capable of efficiently handling these couplings by computing global integrals that encompass all sub-regions.

\subsection{The Basic Framework of the PCNO}\label{sec3.1}
\begin{figure}[h]
\centering
\includegraphics[width=0.9\textwidth]{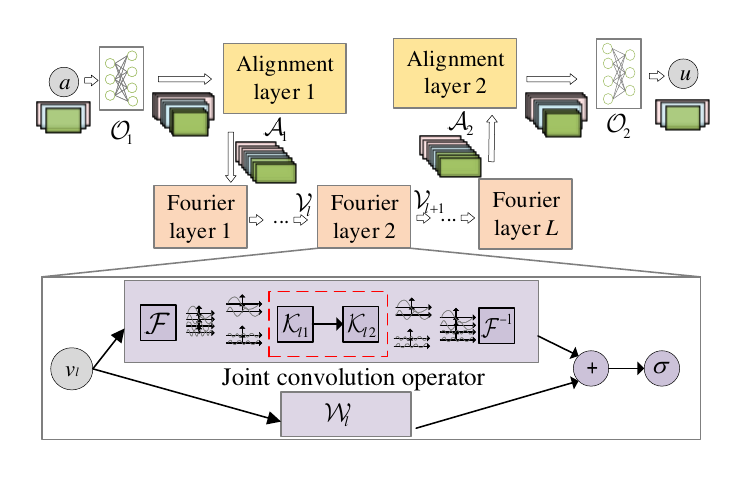}
\caption{The basic framework of the proposed partitioned coupled neural operator. PCNO features two linear layers ${{\mathcal O}_1}:{{\mathbb R}^{{d_a}}} \to {{\mathbb R}^{{d_1}}}$ and ${{\mathcal O}_2}:{{\mathbb R}^{{d_l}}} \to {{\mathbb R}^{{d_u}}}$, two grid alignment layers ${{\mathcal A}_1}$ and ${{\mathcal A}_2}$, and several Fourier layers. The first linear layer lifts the input function $a \in {{\mathbb R}^{{d_a}}}$ from all sub-regions to a higher-dimensional space, while the second projects the output back to the original dimension. ${{\mathcal A}_1}$ 
 ensures consistent input grid sizes across regions to guarantee that Fourier layers compute joint convolutions under the same set of Fourier basis functions.
${{\mathcal A}_2}$ restores the output to the original resolution. The $L$ Fourier layers perform forward and inverse Fourier transforms (${\mathcal F}$ and ${{\mathcal F}^{ - 1}}$), with joint convolution operators ${{\mathcal K}_{l1}}{{\mathcal K}_{l2}}$ used compute Green's function integrals across sub-regions. ${{\mathcal W}_l}$ is defined as a linear transformation in the time domain.}
\label{fig2}
\end{figure}

The framework of the proposed partitioned coupled neural operator is illustrated in Fig.~\ref{fig2}. Unlike traditional Fourier neural operators, the PCNO incorporates grid alignment layers and a Fourier layer featuring a joint convolution operator to address the coupling relationships among multiple sub-regions and compute the Green's function integrals considering inter-sub-region coupling.

\subsection{Discretized Representation of Input Function Spaces}\label{sec3.2}
The input and output function spaces in a partitioned coupled system differ significantly from those in a single-region system. The input and output functions of PCNO are derived from the joint inputs and outputs of all sub-regions, enabling the simultaneous consideration of coupling relationships among multiple regions. Moreover, since the regions influence each other through interface conditions, the grid in each region must integrate the control conditions from all regions. Since the spatial lengths of the regions vary, the grid sizes also differ across regions. However, the grid size in the temporal dimension $t$ remains consistent across all regions.

\begin{figure}[h]
\centering
\includegraphics[width=0.5\textwidth]{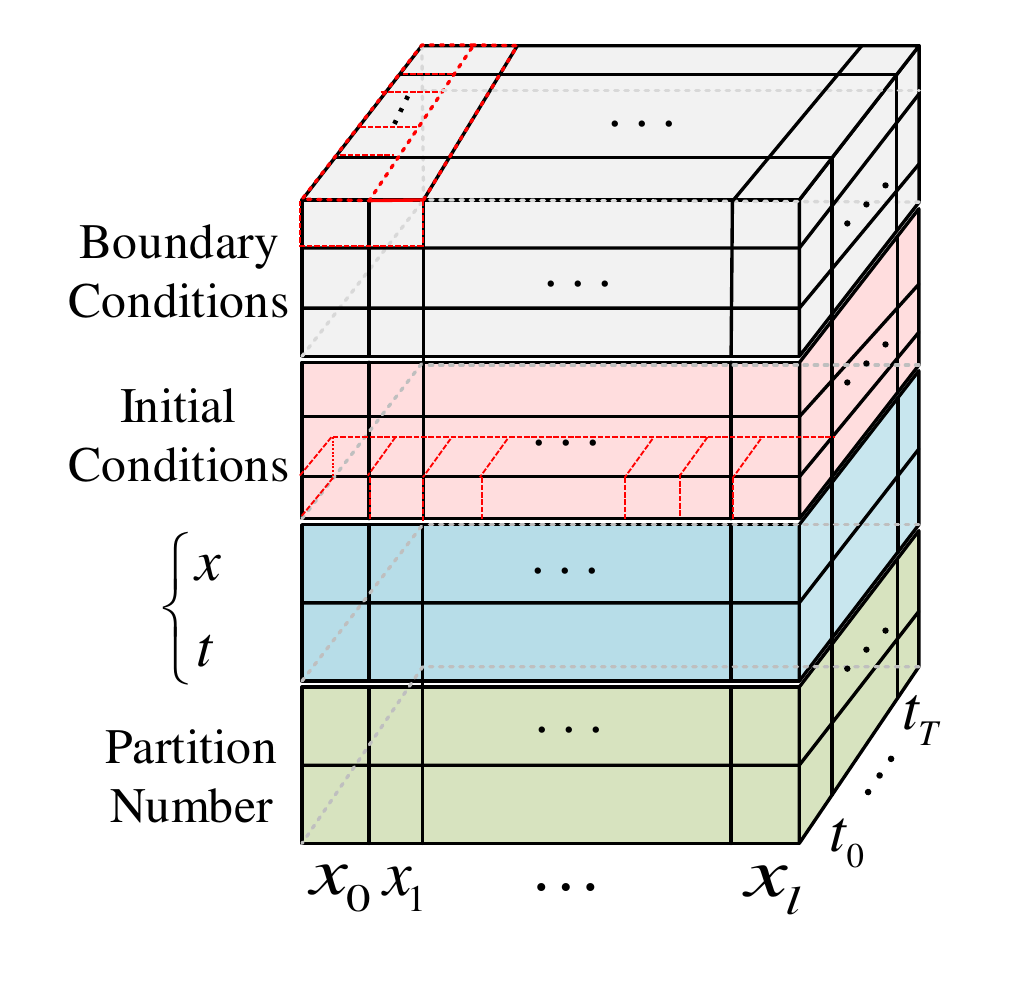}
\caption{Discretized Representation of the Input Function Space for Each Region. Assuming a one-dimensional spatial domain, we construct a two-dimensional discretized grid over time and spatial positions. On this grid, we stack binary-encoded region identifiers along with time, spatial positions, and the boundary and initial conditions, which encompass those of all regions within the network. Notably, boundary conditions are replicated along the spatial dimension, while initial conditions are replicated along the temporal dimension.}
\label{fig3}
\end{figure}

Complex network systems are typically governed by initial and boundary conditions. Assume that the system is divided into $E$ sub-regions, where the evolution of each sub-region is described by a partial differential equation (PDE) in space $x \in {\mathcal D}$ and time $t \in {\mathcal T}$. Each sub-region is discretized on a two-dimensional grid with step sizes $\Delta x$ and $\Delta t$, as illustrated in Fig.~\ref{fig3}. The construction process of the discretized representation of the input function is detailed in the supplementary Section~\ref{sup4}.

\subsection{Grid Alignment Layers}\label{sec3.3}
To address the issue of inconsistent grid sizes across different regions in complex network systems, this paper proposes grid alignment layers. Here, layer   ${{\mathcal A}_1}$ can receive input function encodings from different regions and transforms them into a uniform size, enabling the subsequent Fourier layers to effectively learn the coupling relationships among regions. Since the learning process in the Fourier layer relies on the Discrete Fourier Transform (DFT), it would be challenging to efficiently learn the coupling relationships among sub-regions if different Fourier basis functions were used for different regions. Only when the lengths of the matrices along the transformed dimension are consistent can their Fourier basis functions be aligned, thereby ensuring effective learning of the coupling relationships. At the same time, the initial and boundary conditions of complex network systems are typically non-periodic, while the Fourier Transform performs best when handling periodic signals. In this paper, zero-padding is used at both ends of the time-domain signals to address the issue of non-periodic signals. The specific padding method is provided in Supplementary Section~\ref{sup5}.

Before entering the Fourier layer, the encodings $\{ {a_1}, \cdots {a_E}\}  \in {{\mathbb R}^{{d_1}*{d_t}*{d_x}}}$ for each region are combined to obtain the joint input $a \in {{\mathbb R}^{{d_1}*{d_t}*{d_x}*E}}$. In the grid alignment layer ${{\mathcal A}_2}$, the output encoding is split along the E-dimension into multiple matrices, and the padded dimensions introduced in ${{\mathcal A}_1}$ are directly removed.

\subsection{Fourier Layer with Joint Convolution Operator}\label{sec3.4}

A joint convolution operator is proposed in the Fourier layer to compute global integrals that account for the coupling of sub-regions. Although the PDEs of the system are defined in local regions, the solution of each local region is influenced by other regions. Therefore, the designed neural operator should be global in nature, capable of integrating information from all regions. Inspired by the Green's function \cite{bib21}, the solution of a PDE at any given point is influenced by all regions, and this process can be achieved through integration. When considering the solution at all positions, the integration operation manifests as a convolution \cite{bib17}. According to the convolution theorem, convolution is equivalent to a product in the frequency domain under the Fourier transform. The iterative update structure of the Fourier layer ${v_l} \to {v_{l + 1}}$ is as follows:

\begin{equation}
\begin{matrix}
   {{v_{l + 1}}(x,t) = \sigma ({{\mathcal W}_l}{v_l}(x,t) + \left( {{{\mathcal K}_{l2}}({{\mathcal K}_{l1}}{v_l})} \right)(x,t))} & {\forall (x,t) \in ({\mathcal D},{\mathcal T})}  \\ 
\end{matrix} 
\label{eq3}
\end{equation}
here, the input is ${v_l} \in {{\mathbb R}^{{d_l}*{d_t}*{d_x}*E}}$ and the output is ${v_{l + 1}} \in {{\mathbb R}^{{d_{l + 1}}*{d_t}*{d_x}*E}}$. ${{\mathcal W}_l} \in {{\mathbb R}^{{d_{l + 1}}*{d_l}}}$ is a linear transformation parameterized by the convolutional neural network. ${{\mathcal K}_{l1}}$ and ${{\mathcal K}_{l2}}$  together form the joint convolution operator, which is key for implementing multi-region global integrals in the Fourier domain. $\sigma$ represents the activation function.

The joint convolution operator is defined as follows:

\begin{equation}
\begin{matrix}
   {\left( {{{\mathcal K}_{l2}}({{\mathcal K}_{l1}}{v_l})} \right)(x,t) = {{\mathcal F}^{ - 1}}({{\mathcal K}_{l2}} \cdot {{\mathcal K}_{l1}} \cdot {\mathcal F}{v_l})(x,t)} & {\forall (x,t) \in ({\mathcal D},{\mathcal T})}  \\ 
\end{matrix} 
\label{eq4}
\end{equation}
here, the input ${v_l} \in {{\mathbb R}^{{d_l}*{d_t}*{d_x}*E}}$ undergoes a real-valued fast Fourier transform (FFT) in the temporal and spatial dimensions, resulting in ${\mathcal F}{v_l} \in {{\Bbb C}^{{d_l}*{d_t}*{d_x}*E}}$. Subsequently, similar to FNO-2D \cite{bib17}, we apply a cutoff in the Fourier frequency components to reduce the sizes of the convolution operators ${{\mathcal K}_{l1}}$ and ${{\mathcal K}_{l2}}$. In the temporal dimension, we truncate the highest and the lowest ${Z_{t,\max }}$ frequency components, while in the spatial dimension, we only truncate the lowest ${Z_{x,\max }}$ frequency components. Through these operations, we obtain the high-frequency components ${\mathcal F}{v_l}^{high} \in {{\Bbb C}^{{d_l}*{d_{{Z_{t,\max }}}}*{d_{{Z_{x,\max }}}}*E}}$ and the low-frequency components ${\mathcal F}{v_l}^{low} \in {{\Bbb C}^{{d_l}*{d_{{Z_{t,\max }}}}*{d_{{Z_{x,\max }}}}*E}}$. In the frequency domain, ${{\mathcal K}_{l1}}$ and ${{\mathcal K}_{l2}}$ are each parameterized separately by the low- and high-frequency components, with each having distinct complex-valued parameters for the low-frequency and high-frequency components. Finally, the results of ${{\mathcal K}_{l2}}^{high} \cdot {{\mathcal K}_{l1}}^{high} \cdot {\mathcal F}{v_l}^{high}$ and ${{\mathcal K}_{l2}}^{low} \cdot {{\mathcal K}_{l1}}^{low} \cdot {\mathcal F}{v_l}^{low}$  are concatenated, and an inverse fast Fourier transform (IFFT) is applied.

The convolution operations for both high- and low-frequency components are consistent. For the low-frequency components ${\mathcal F}{v_l}^{low}$, the convolution operator ${{\mathcal K}_{l1}}^{low}$ is defined as a complex-valued parameter ${{\Bbb C}^{{d_{{Z_{t,\max }}}}*{d_{{Z_{x,\max }}}}*E*E}}$, while ${{\mathcal K}_{l2}}^{low} \in {{\Bbb C}^{{d_{l + 1}}*{d_l}*E}}$. The operation ${{\mathcal K}_{l1}}^{low} \cdot {\mathcal F}{v_l}^{low}$ is a linear transformation applied in the $E$ dimension, where it assigns weights to different regions and performs convolution operations across temporal and spatial dimensions. Next, the operation ${{\mathcal K}_{l2}}^{low} \cdot {{\mathcal K}_{l1}}^{low} \cdot {\mathcal F}{v_l}^{low}$ applies the weight to the $E$ dimension, integrating information from different channels. Since the connection between regions varies, ${{\mathcal K}_{l2}}^{low}$ incorporates the regional dimensionality $E$, ensuring that each region can independently integrate channel information, thereby improving accuracy.

The ${v_l}$ achieves cross-domain interaction through convolutions with ${{\mathcal K}_{l1}}$ and ${{\mathcal K}_{l2}}$. Although it is possible to perform the Fourier transform across the ${d_t}$, ${d_x}$ and $E$ dimensions simultaneously and complete the global convolution using a single operator ${\mathcal K} \in {{\Bbb C}^{{d_{l + 1}}*{d_l}*{d_{{Z_{t,\max }}}}*{d_{{Z_{x,\max }}}}*E}}$, the signal ${v_l}$ may be discontinuous in the $E$ dimension. As a result, the Fourier transform may include excessive high-frequency components, leading to unwanted high-frequency noise in the transformed frequency spectrum. The experimental results can be found in the Table~\ref{tab3} under the method PCNO-3D.

However, since ${v_l}$ is defined in the functional space of the temporal and spatial dimensions, it maintains continuity between discrete points within this space. This implies that only by performing the Fourier transform in both the temporal and spatial dimensions can the correct spectrum be obtained.

\subsection{Physics-Informed Neural Operator Learning}\label{sec3.5}
Under the control of the input initial and boundary conditions $a$, the complex network system can solve the PDEs for each subregion to obtain the physical field $u$ in each region. The goal of this section is to learn a neural operator ${{\mathcal G}_\theta }$ with parameter $\theta $, which will serve as an approximation to the true solver ${\mathcal G}$.

Although the neural operator is defined to learn the mapping between two infinite-dimensional function spaces, the practical operation involves training on discrete data pairs $\{ {a_j},{u_j}\} _{j = 1}^N$ obtained from discretized inputs and outputs. To address the challenge of accurately approximating the true solver with low-resolution observational data, this paper integrates high-fidelity physical information with data to jointly train the neural operator ${{\mathcal G}_\theta }$. In addition, the PDE loss function serves as a soft constraint that directly penalizes errors in the solution space. Moreover, the PDE loss function for complex network systems is a complex soft constraint, making direct optimization challenging. Therefore, it is necessary to combine data-driven methods to efficiently optimize it through the enforcement of hard constraints. The loss function for the observational data is defined as:

\begin{equation}
{{\mathcal L}_{data}} = \frac{1}{E}\sum\limits_{e \in E} {{{\left\| {{u_e} - {{\mathcal G}_{\theta ,e}}(a)} \right\|}_{{\Gamma ^2}({\mathcal T},{\mathcal D})}}} 
\label{eq5}
\end{equation}
here, ${u_e}$ represents the known data in region $e$, and ${{\mathcal G}_{\theta ,e}}(a)$ represents the output of the neural operator in region $e$. $({\mathcal T},{\mathcal D})$ is the value ranges of $t$ and $x$. $|| \cdot |{|_{{\Gamma ^2}}}$ denotes the root mean square error

In addition, the embedding of physical information is achieved by minimizing the loss function. The PDE loss function consists of three parts: equation loss ${{\mathcal L}_{EQ}}$, initial condition loss ${{\mathcal L}_I}$, and boundary condition loss ${{\mathcal L}_B}$. Among them, $\alpha $, $\beta $ and $\delta $ are weight coefficients. It is worth noting that in complex network systems, certain nodes are subject to known control inputs, while junction nodes are governed by conservation laws, as illustrated in the supplementary Section~\ref{sup1.2}. The boundary condition losses for these nodes are computed separately.

\begin{equation}
{{\mathcal L}_{PDE}} = \alpha {{\mathcal L}_{EQ}} + \beta {{\mathcal L}_I} + \delta {{\mathcal L}_B}
\label{eq6}
\end{equation}

To compute the PDE loss, it is necessary to calculate the derivative terms $du/dx$ and $du/dt$. Generally, numerical differentiation, neural network automatic differentiation \cite{bib43}, or functional differentiation in the Fourier domain \cite{bib39} can be used to calculate these derivative terms.

\subsection{Configuration of PCNO for Natural Gas Network Simulation}\label{sec2.1}

The natural gas network is governed by time-varying boundary conditions, including the pressure at the source node of pipeline 1 and the flow rates at the demand nodes of pipelines 2 and 3. PCNO aims to predict the flow and pressure distribution across all pipelines within the time span of boundary control. The network’s inlet pressure is fixed at 0.3 MPa throughout the process, while flow rates boundary conditions change dynamically over a 24-hour period. The flow boundaries are controlled by randomly generated square waves. The values of the square waves are adjusted randomly between 0.5 kg/s and 2 kg/s, with the number of adjustments ranging from 0 to 24 times. This setting ensures the simulation captures extreme operational conditions in the gas network.

The PCNO is trained by combining the physics-informed loss ${{\mathcal L}_{PDE}}$ (see Eq.~\ref{eq8}) and the observation data loss ${{\mathcal L}_{data}}$ (see Eq.\ref{eq7}).  The training data are obtained from simulation results generated by the Pipeline Studio software. Since the boundary conditions in the simulation process cannot be adjusted instantaneously, there are subtle differences between the boundary conditions of the simulation system and those of the PDE. The PCNO contains four Fourier layers, with the dimension of the intermediate space set to ${d_l} = {d_{l + 1}} = 64$. The Fourier cutoff frequency is set to ${Z_{t,\max }} = {Z_{x,\max }} = 25$. Additionally, the spatial fill ratio for the grid alignment layer is set to ${r_x} = 0.001$, and the temporal fill ratio to ${r_t} = 0$.

\bmhead{Acknowledgements}

This work was supported by the National Natural Science Foundation of China (No. 62133015), the 2024 Jiangsu Province Graduate Research and Innovation Program (No. KYCX24\_2770), and the 2024 Graduate Innovation Program of China University of Mining and Technology (No. 2024WLKXJ084).

\bmhead{Data Availability Statement}

\bmhead{Code Availability statement}

\bmhead{Author contributions}

\bmhead{Competing interests}

The authors declare no competing financial or non-financial interests.

\section*{Declarations}

Some journals require declarations to be submitted in a standardised format. Please check the Instructions for Authors of the journal to which you are submitting to see if you need to complete this section. If yes, your manuscript must contain the following sections under the heading `Declarations':

\begin{itemize}
\item Funding
\item Conflict of interest/Competing interests (check journal-specific guidelines for which heading to use)
\item Ethics approval and consent to participate
\item Consent for publication
\item Data availability 
\item Materials availability
\item Code availability 
\item Author contribution
\end{itemize}

\noindent
If any of the sections are not relevant to your manuscript, please include the heading and write `Not applicable' for that section. 

\bigskip
\begin{flushleft}%
Editorial Policies for:

\bigskip\noindent
Springer journals and proceedings: \url{https://www.springer.com/gp/editorial-policies}

\bigskip\noindent
Nature Portfolio journals: \url{https://www.nature.com/nature-research/editorial-policies}

\bigskip\noindent
\textit{Scientific Reports}: \url{https://www.nature.com/srep/journal-policies/editorial-policies}

\bigskip\noindent
BMC journals: \url{https://www.biomedcentral.com/getpublished/editorial-policies}
\end{flushleft}


\bibliography{bibliography}

\newpage
\begin{appendices}

\end{appendices}

\bmhead{Supplementary information}

\begin{table}[h]
  \centering
  \caption{Parameters of Natural Gas Pipelines}
    \begin{tabular}{cccccccc}
    \toprule
    \multicolumn{1}{r}{} & \multicolumn{1}{c}{Diameter(m)} & \multicolumn{1}{c}{Length(m)} & \multicolumn{1}{c}{$\lambda$} & \multicolumn{1}{c}{Z} & \multicolumn{1}{c}{R} & \multicolumn{1}{c}{T(K)} & \multicolumn{1}{c}{E} \\
    \midrule
    Pipe 1 & 0.5   & 60000 & 0.011851315 & 1     & 288.0937 & 288.15 & 1 \\
    Pipe 2 & 0.6   & 50000 & 0.011152515 & 1     & 288.0937 & 288.15 & 1 \\
    Pipe 3 & 0.7   & 40000 & 0.010593932 & 1     & 288.0937 & 288.15 & 1 \\
    \bottomrule
    \end{tabular}%
  \label{tab4}%
\end{table}%

\begin{table}[h]
  \centering
  \caption{Coefficient Table for Data and PDE Loss Functions}
    \begin{tabular}{p{8em}p{8em}p{8em}p{8em}}
    \toprule
          & 0     & 1 & 2 \\
    \midrule
      $\alpha$    & 0.2   & 2.5 & 30 \\
      $\beta$   & 0.1   & 2.5 & 30 \\
      $\delta$    & 0.2   & $ZRT$   & 1 \\
    \bottomrule
    \end{tabular}%
  \label{tab5}%
\end{table}%

\section{Natural Gas Network Model}\label{sup1}
\subsection{Mathematical Model of Natural Gas Pipeline}\label{sup1.1}
The one-dimensional compressible heat-conducting transient flow model for natural gas pipelines includes the continuity equation ${f_1}$, the momentum equation ${f_2}$, and the energy equation  ${f_3}$ \cite{bib53}, as shown in Eq.~\ref{eq19}-\ref{eq21}. The continuity and momentum equations are derived from the laws of mass conservation and momentum conservation, respectively, and ${f_2}$ is specific cases of the Navier-Stokes (NS) equations under certain assumptions. These equations describe the flow rate, pressure, density, and velocity at various positions and times along the pipeline, while the energy equation is used to describe states such as temperature and enthalpy.

\begin{equation}
    \frac{{\partial \rho }}{{\partial t}} + \frac{{\partial (\rho v)}}{{\partial x}} = 0
\label{eq19}
\end{equation}

\begin{equation}
 \frac{{\partial (\rho v)}}{{\partial t}} + \frac{{\partial (\rho {v^2})}}{{\partial x}} + \frac{{\partial (p)}}{{\partial x}} =  - \frac{\lambda }{2}\frac{{\rho v|v|}}{D} - g\rho A{\text{sin}}\theta 
\label{eq20}
\end{equation}

\begin{equation}
    \frac{\partial }{{\partial t}}\left[ {\rho \left( {u + \frac{{{v^2}}}{2} + gz} \right)} \right] + \frac{\partial }{{\partial x}}\left[ {(\rho v)\left( {h + \frac{{{v^2}}}{2} + gz} \right)} \right] =  - \frac{{\partial {Q_q}}}{{\partial x}}\rho v
\label{eq21}
\end{equation}
where, $t \in {\mathcal T}$ and $x \in {\mathcal D}$ represent time (in seconds) and position (in meters) along the pipeline, respectively. $\rho$, $v$, $P$ denote the natural gas density $\mathrm{(kg/m^3)}$, velocity (m/s), and pressure (Pa), respectively. $\lambda $, $A$, $D$, $z$ are the pipe friction factor, cross-sectional area $(\mathrm{m^2})$, diameter $\mathrm{(m)}$, and elevation, respectively. $\theta $ is the inclination angle between the horizontal plane and the pipeline. $g$, $u$, $h$, ${Q_q}$ represent gravitational acceleration $(\mathrm{m/s^2})$, specific internal energy $\mathrm{(J/kg)}$, specific enthalpy (J/kg), and the heat exchange per unit mass between the gas and its surrounding (J/kg) environment, respectively. 

The mass flow rate of natural gas is defined as $M = \rho vA$, with units in kilograms per second (kg/s). Additionally, the ideal gas equation of state describes the relationship between pressure, gas composition, temperature, and density. The ideal gas law is employed in this paper \cite{bib54}.

\begin{equation}
    P = ZRT\rho 
\label{eq22}
\end{equation}
where, $Z$ represents the compressibility factor, $R$ denotes the natural gas constant, and $T$ is the temperature.

The friction factor is derived from the Weymouth equation, with $E$ being the efficiency coefficient, as referenced in Eq.~\ref{eq23}.

\begin{equation}
    \sqrt {1/f}  = 11.19{D^{0.167}}E
\label{eq23}
\end{equation}

When the flow velocity is much lower than the speed of sound, the convective term in the momentum equation $\partial (\rho {v^2})/\partial x$ approaches zero and is often neglected in engineering calculations for simplicity \cite{bib59}. The pipeline inclination is related to the terrain elevation difference, but here we assume that the inclination of the natural gas pipeline is zero. Consequently, the momentum conservation equation is simplified to Eq.~\ref{eq24}. Additionally, it is assumed that the natural gas transmission is isothermal, meaning that the energy eq.~\ref{eq21} is ignored \cite{bib51}.

\begin{equation}
\frac{{\partial (\rho v)}}{{\partial t}} + \frac{{\partial (p)}}{{\partial x}} + \frac{\lambda }{2}\frac{{\rho v|v|}}{D} = 0
\label{eq24}
\end{equation}

\subsection{Pipeline Network Constraints}\label{sup1.2}
In a natural gas pipeline network, nodes can be classified as supply nodes ${N_s}$, located at the natural gas supply end, junction nodes ${N_J}$ where pipelines intersect, and demand nodes ${N_D}$ at the consumption end. It is important to note that junction nodes may coincide with either supply or demand nodes. The junction node ${N_J}$ satisfies the mass conservation condition, meaning that the mass flow rate entering the node is equal to the mass flow rate exiting the node, as shown in Eq.~\ref{eq25}.

\begin{equation}
\sum\limits_{i \in in} {{m_{j,i}}}  = \sum\limits_{o \in out} {{m_{j,o}}}  + \sum\limits_{{N_d} \in {N_D}} {{m_{j,{N_d}}}} 
\label{eq25}
\end{equation}
where ${m_{j,i}}$ is the mass flow rate entering node $j$ from another node $i$, ${m_{j,o}}$ is the mass flow rate exiting node $j$ to another node $o$, and ${m_{j,{N_d}}}$ is the mass flow rate of natural gas consumed by the user at node $j$.

Additionally, at any given time, the gas pressure along the boundary of the pipeline connected to the junction node remains equal, as expressed in Eq.~\ref{eq26}.

\begin{equation}
{P_{i,t}} = {P_{o,t}} = {P_{{N_d},t}} =  \ldots 
\label{eq26}
\end{equation}

\subsection{Initial and Boundary Conditions}\label{sup1.3}
The specification of initial and boundary conditions ensures that the partial differential equation system yields a unique and physically reasonable solution. Initial conditions are necessary to uniquely determine the instantaneous state of the fluid and are typically set to either the initial steady-state flow or the final state from the previous time step. In this study, we assume steady-state initial conditions, whereby the time-dependent partial derivative terms in Eq.~\ref{eq19} and Eq.~\ref{eq20} become zero.

\begin{equation}
\begin{cases}
\frac{\partial(\rho\nu)}{\partial x}=0\\\frac{\partial(p)}{\partial x}+\frac{\lambda}{2}\frac{\rho\nu|\nu|}{D}=0
\end{cases}
\label{eq27}
\end{equation}
here, $\sigma $ is a coefficient, and $\Delta t$, $\Delta x$ represent the time step and spatial step, respectively.

Additionally, to define the boundary conditions in system, it is common to set known pressures or flow rates at the pipeline network’s inlet and outlet, along with the boundary conditions for system temperature as it varies over time \cite{bib53}. In this study, the known boundary conditions include the inlet pressure and the outlet consumption flow rate, with the system temperature assumed to be constant.

\section{Application in the Natural Gas Pipeline System}\label{sup2}
This section uses the natural gas pipeline system as an example to further illustrate the practical application of the proposed PCNO, with the system parameters provided in Supplementary Table~\ref{tab4}. First, the input-output format of the natural gas pipeline system model is outlined, followed by a discussion of how the physical information is embedded.

\subsection{Multi-region Representation of the System's Input Function}\label{sec4.1}
Similar to other networks, the natural gas pipeline system operates under the control of initial and boundary conditions. Initial conditions are essential for determining the instantaneous state of the fluid, and are typically set as the initial steady flow or the terminal state of the previous time step. Additionally, to determine the system's  boundary state, known pressure or flow rates are typically assigned at the inlet and outlet of the pipeline, along with boundary conditions that describe the variation of system temperature over time \cite{bib53}. In this study, the known boundary conditions are the inlet pressure and the consumption flow at the outlet, while the system temperature is assumed to remain constant. Furthermore, the initial condition is set as a steady-state flow, which is determined by the boundary conditions at time $t = 0$. This is treated as an implicit condition and does not need to be explicitly expressed in the input. As a result, the input function of the pipeline system only includes regional encoding, grid positions $(x,t)$, and boundary conditions.

Specifically, the multi-region input of the system can be represented as $a = \{ {a_1} \cdots {a_E}\} $, where the input function for the $e$-th region is ${a_e} = (e,x,t,M_t^{{N_D}},P_t^{{N_S}}) \in {{\mathbb R}^{{d_a}*{d_t}*{d_{x,e}}}}$ and $\{x,t\} \in \{\mathcal{T},\mathcal{D}\}$. $M_t^{{N_D}}$ and $P_t^{{N_S}}$ represent the flow boundary conditions at demand nodes and the pressure boundary conditions at source nodes, respectively. The output function represents the pressure and flow rate of the system at any position and time, denoted as $u = \{ {u_1} \cdots {u_E}\} $, where ${u_e} = ({M_{x,t}},{P_{x,t}})$ and $\{x,t\} \in \{\mathcal{T},\mathcal{D}\}$.

\subsection{Multi-region Representation of the System's Input Function}\label{sec4.2}
Since the solver outputs flow rates and pressures at different magnitudes, the loss for the observational data is defined as follows.

\begin{equation}
{{\mathcal L}_{data}} = \frac{{{\gamma _1}}}{E}\sum\limits_{e \in E} {{{\left\| {{M_e} - {M_{{u_\theta },e}}} \right\|}_{{\Gamma ^2}({\mathcal T},{\mathcal D})}}}  + \frac{{{\gamma _2}}}{E}\sum\limits_{e \in E} {{{\left\| {{P_e} - {P_{{u_\theta },e}}} \right\|}_{{\Gamma ^2}({\mathcal T},{\mathcal D})}}} 
\label{eq7}
\end{equation}
where, ${\gamma _1}$ and ${\gamma _2}$ are weighting coefficients, ${M_e}$ and ${P_e}$ represent the true flow rate and pressure in region $e$. ${M_{{u_\theta },e}}$ and ${P_{{u_\theta },e}}$ represent the flow rate and pressure output by the solver in region $e$.

The model of the natural gas pipeline system is provided in Supplementary Information Section~\ref{sup1}. Its PDE loss function consists of four parts: the equation loss (see Eq.~\ref{eq19} and Eq.~\ref{eq24}), the initial conditions (see Eq.~\ref{eq27}), the boundary conditions (see Eq.~\ref{eq25} and Eq.~\ref{eq26}). It is worth noting that the control equations of the natural gas network system consist of ${f_1}$ and ${f_2}$, with boundary conditions covering two physical quantities: flow and pressure. The losses for each physical quantity are calculated separately at known control nodes and junctions. 

\begin{equation}
{{\mathcal L}_{PDE}} = \alpha_0 {{\mathcal L}_{EQ}} + \beta_0 {{\mathcal L}_I} + \delta_0 {{\mathcal L}_B}
\label{eq8}
\end{equation}

\begin{equation}
{{\mathcal L}_{EQ}} = \frac{1}{E}\sum\limits_{e \in E} {{\alpha _1}{{\left\| {{f_{1,{u_\theta },e}}} \right\|}_{{\Gamma ^2}({\mathcal T},{\mathcal D})}} + {\alpha _2}{{\left\| {{f_{2,{u_\theta },e}}} \right\|}_{{\Gamma ^2}({\mathcal T},{\mathcal D})}}} 
\label{eq9}
\end{equation}

\begin{equation}
{\mathcal{L}_I} = \frac{1}{E}\sum\limits_{e \in E} {{\beta _1}{{\left\| {f_{_{1,{u_\theta },e}}^{\prime}} \right\|}_{{\Gamma ^2}(0,\mathcal{D})}} + {\beta _2}{{\left\| {f_{_{2,{u_\theta },e}}^{\prime}} \right\|}_{{\Gamma ^2}(0,\mathcal{D})}}} 
\label{eq10}
\end{equation}

\begin{equation}
{{\mathcal L}_B} = {\delta _1}{B_M} + {\delta _2}{B_P}
\label{eq11}
\end{equation}

\begin{equation}
\begin{aligned}
      {B_M} = &\frac{1}{{|{N_D}|}}\sum\limits_{{N_d} \in {N_D}} {{{\left\| {{M^{{N_d}}} - M_{{u_\theta }}^{out,{N_d}}} \right\|}_{{\Gamma ^2}({\mathcal T},\partial {\mathcal D} \in {N_D})}}}   \\   
  &  + \frac{1}{{|{N_J}|}}\sum\limits_{{N_j} \in {N_J}} {{{\left\| {M_{{u_\theta }}^{in,{N_j}} - M_{{u_\theta }}^{out,{N_j}}} \right\|}_{{\Gamma ^2}({\mathcal T},\partial {\mathcal D} \in {N_J})}}} 
\end{aligned} 
\label{eq12}
\end{equation}

\begin{equation}
\begin{aligned}
  {B_P} &= \frac{1}{{|{N_S}|}}\sum\limits_{{N_s} \in {N_S}} {\frac{1}{{|{N_s}|}}\sum\limits_{m \in {N_s}} {{{\left\| {P_{_{{u_\theta },m}}^{{N_s}} - {P^{{N_s}}}} \right\|}_{{\Gamma ^2}({\mathcal T},\partial {\mathcal D} \in {N_S})}}} }   \\
  &  + \frac{1}{{|{N_J}|}}\sum\limits_{{N_j} \in {N_J}} {\frac{1}{{C_{|{N_j}|}^2}}\sum\limits_{{j_1},{j_2} \in {N_j},{j_1} \ne {j_2}} {{{\left\| {P_{_{{u_\theta },{j_1}}}^{{N_j}} - P_{_{{u_\theta },{j_2}}}^{{N_j}}} \right\|}_{{\Gamma ^2}({\mathcal T},\partial {\mathcal D} \in {N_J})}}} }  
\end{aligned}
\label{eq13}
\end{equation}
where, $\alpha ,\beta$ and $\delta $ are weighting coefficients, with their values provided in Supplementary Table~\ref{tab5}. ${f_{1,{u_\theta },e}}$ represents the result of computing equation ${f_1}$ for the operator output ${u_\theta }$ in region $e$. $f_1^{\prime}$ and $f_2^{\prime}$  are the PDE under the initial steady-state conditions. $M_{{u_\theta }}^{in,*}$ and $M_{{u_\theta }}^{out,*}$ represent the inlet and outlet flow rates of the output ${u_\theta }$ at a given node. $P_{_{{u_{\theta ,a}}}}^*$ is the boundary pressure of the output ${u_\theta }$ in region $a$, connected to a given node. ${M^{{N_d}}}$ and ${P^{{N_s}}}$ denote the known flow rate at node ${N_d}$ and the known pressure at node ${N_s}$, respectively. $E$, $|{N_J}|$, $|{N_S}|$, $|{N_D}|$ represent the number of regions, junction nodes, source nodes, and demand nodes, respectively. $|{N_s}|$, $|{N_j}|$ denote the number of regions connected to the $s$-th source node and  $j$-th junction nodes ${N_j}$, respectively; $m$ represents the region connected to the $s$-th source node. ${j_1}$ and ${j_2}$ represent the regions connected to the $j$-th junction node. $\left( {{\mathcal T},\partial {\mathcal D} \in {N_D}} \right)$, $\left( {{\mathcal T},\partial {\mathcal D} \in {N_S}} \right)$ and $\left( {{\mathcal T},\partial {\mathcal D} \in {N_J}} \right)$ represent the boundary sets for the demand nodes, source nodes, and the junction nodes, respectively.

The natural gas network has a high-resolution uniform grid and a non-periodic target function, making numerical differentiation methods more advantageous for quickly and accurately calculating its partial derivatives. In this paper, the four-point implicit finite difference method \cite{bib60} is used to discretize the continuity equation ${f_1}$ and the momentum equation ${f_2}$. The discrete format is shown in Eq.~\ref{eq14}.

\begin{equation}
\begin{aligned}
    \begin{cases}
    &f\left(x,t\right)=\frac{\sigma}{2}\Big(f_x^{t+1}+f_{x+1}^{t+1}\Big)+\frac{1-\sigma}{2}\Big(f_x^t+f_{x+1}^t\Big)\\
    &\frac{\partial f}{\partial t}=\frac{\Big(f_{x+1}^{t+1}-f_{x+1}^t\Big)+\Big(f_x^{t+1}-f_x^t\Big)}{2\Delta t}\\
    &\frac{\partial f}{\partial x}=\sigma\frac{f_{x+1}^{t+1}-f_{x}^{t+1}}{\Delta x}+\left(1-\sigma\right)\frac{f_{x+1}^{t}-f_{x}^{t}}{\Delta x}
    \end{cases}
\end{aligned}
\label{eq14}
\end{equation}

\begin{equation}  
\begin{aligned}
   &P_x^{t + 1} - P_x^t + P_{x + 1}^{t + 1} - P_{x + 1}^t +    \\
  &\frac{{ZRT\Delta t}}{{\Delta x}}\left[ {\sigma \left( {\rho v_{x + 1}^{t + 1} - \rho v_x^{t + 1}} \right) + \left( {1 - \sigma } \right)\left( {\rho v_{x + 1}^t - \rho v_x^t} \right)} \right] = 0 \
\end{aligned}
\label{eq15}
\end{equation}

\begin{equation}
\begin{aligned}
  & \frac{{\Delta x}}{{\Delta t}}\left( {\rho v_x^{t + 1} - \rho v_x^t + \rho v_{x + 1}^{t + 1} - \rho v_{x + 1}^t} \right) +   \\ 
  & \left[ {\sigma \left( {P_{x + 1}^{t + 1} - P_x^{t + 1}} \right) + \left( {1 - \sigma } \right)\left( {P_{x + 1}^t - P_x^t} \right)} \right] +   \\ 
  & \frac{{\lambda \Delta x}}{{2D}}\left( {\frac{\sigma }{2}(\rho v|v|_x^{t + 1} + \rho v|v|_{x + 1}^{t + 1}) + \frac{{1 - \sigma }}{2}(\rho v|v|_x^t + \rho v|v|_{x + 1}^t)} \right) = 0 
\end{aligned}
\label{eq16}
\end{equation}

As a result, the simplified two equations Eq.~\ref{eq19} and Eq.~\ref{eq20} are discretized using eq.~\ref{eq14}. Furthermore, to reduce the magnitude differences between the loss functions, these two equations are adjusted as follows:

It is important to note that the four-point difference format does not exist under initial conditions Eq.~\ref{eq27}, so the discretized forms of ${f_1^{\prime}}$ and ${f_2^{\prime}}$ are as follows:
\begin{equation}
    \frac{{ZRT\Delta t}}{{\Delta x}}\left[ {\left( {\rho v_{x + 1}^t - \rho v_x^t} \right)} \right] = 0
\label{eq17}
\end{equation}

\begin{equation}
    \left( {P_{x + 1}^t - P_x^t} \right) + \frac{{\lambda \Delta x}}{{4D}}\left( {\rho v|v|_x^t + \rho v|v|_{x + 1}^t} \right) = 0
\label{eq18}
\end{equation}

\section{Baseline}\label{sup3}
(1) PCNO-C: it is a variant of PCNO in which all regions share the same parameters. It is used to test whether assigning equal weights to the information from different regions affects the accuracy of the operator. PCNO-C includes a grid alignment layer, and ${{\mathcal K}_{l1}}$ is configured similarly to PCNO, while ${{\mathcal K}_{l2}}$ is replaced by ${{\mathcal K}_{l2}} \in {{\Bbb C}^{{d_{l + 1}}*{d_l}}}$.

(2) PCNO-3D: it is a variant of PCNO that performs Fourier transforms across the time, spatial, and regional dimensions simultaneously. It is designed to evaluate the impact of including regional dimensions in the Fourier transform on the accuracy of the operator. PCNO-3D contains a grid alignment layer but removes ${{\mathcal K}_{l1}}$ and ${{\mathcal K}_{l2}}$, replacing them with ${\mathcal K} \in {{\Bbb C}^{{d_{l + 1}}*{d_l}*{d_{{Z_{t,\max }}}}*{d_{{Z_{x,\max }}}}*{d_{{Z_{e,\max }}}}}}$. We set the truncation frequency for the regional dimension to ${d_{{z_{e,\max }}}} = 3$, which ensures that the operator can capture all necessary regional information from the three gas pipelines. This configuration ensures that the operator can retrieve the information from each region. Only by ensuring sufficient sampling of the frequency in the regional dimension can we determine the impact of including the regional dimension in the Fourier transform.

(3) FNO-2D: FNO-2D, as proposed in \cite{bib17}, is the standard Fourier neural operator capable of performing convolutions within individual regions but unable to capture interactions across regions. It is equivalent to removing ${{\mathcal K}_{l1}}$ and ${{\mathcal K}_{l2}}$, replacing them with ${\mathcal K} \in {{\Bbb C}^{{d_{l + 1}}*{d_l}*{d_{{Z_{t,\max }}}}*{d_{{Z_{x,\max }}}}}}$. Since we set the grid alignment parameters ${r_x} = {r_t} = 0$, the grid sizes across regions are inconsistent. Although input functions from all three regions are fed into the operator simultaneously, they are processed independently, meaning each region’s convolution uses different parameters.

(4) FNO-3D: We further extended FNO-3D in \cite{bib17} by incorporating a grid alignment layer, allowing regional dimensions to participate in the Fourier transform. However, compared to PCNO-3D, this modification increases the complexity of hyperparameters. Due to inconsistent grid sizes across regions, grid alignment remains necessary. FNO-3D applies Fourier transforms across the time, spatial, and regional dimensions, and performs truncation based on the maximum frequency range. Truncation occurs at the highest and lowest frequencies ${Z_{t,\max }}$ and ${Z_{x,\max }}$ for the time and spatial dimensions, and at the lowest frequency ${d_{{z_{e,\max }}}} = 3$ for the regional dimension. Thus, ${\mathcal K} \in {{\Bbb C}^{{d_{l + 1}}*{d_l}*{d_{{Z_{t,\max }}}}*{d_{{Z_{x,\max }}}}*{d_{{Z_{e,\max }}}}}}$ requires four frequency parameters, two more than PCNO-3D.

\section{The construction process of the discretized representation of the input function}\label{sup4}

\begin{figure}[h]
\centering
\includegraphics[width=0.5\textwidth]{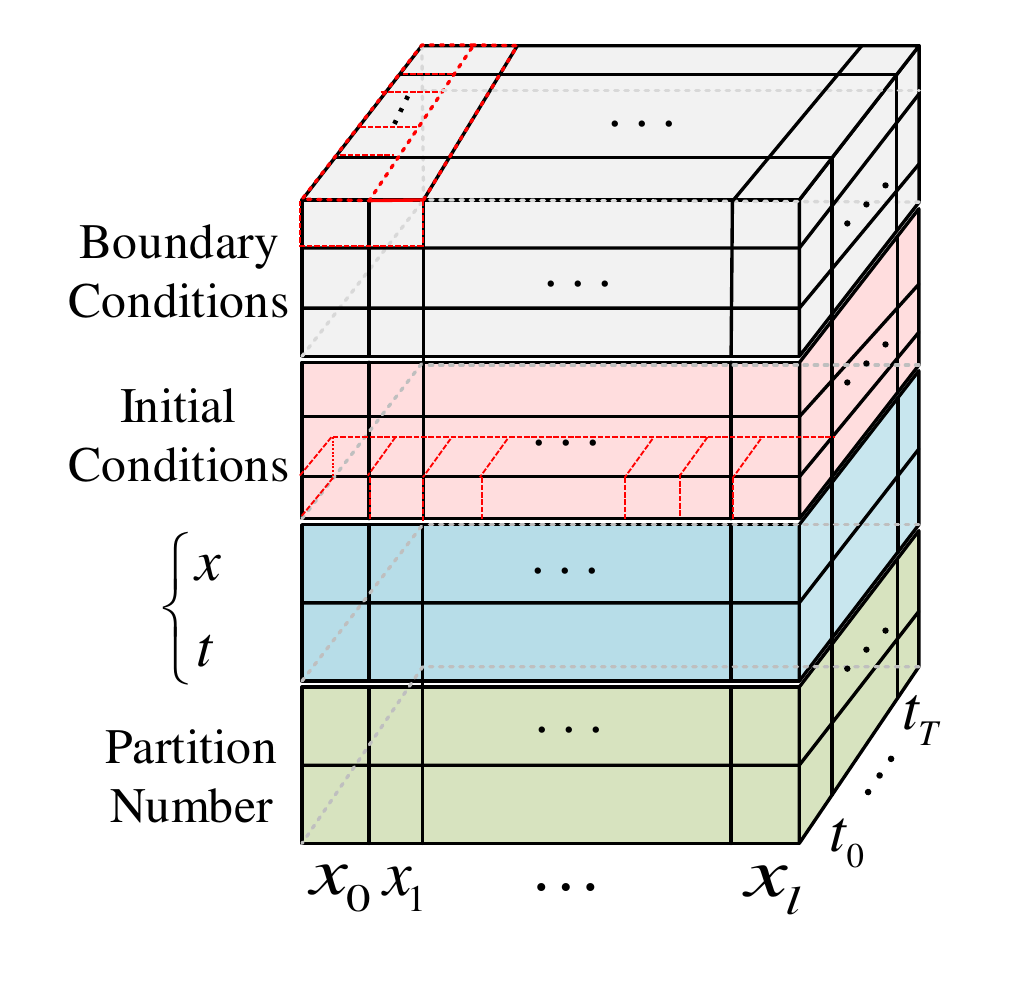}
\caption{Discretized Representation of the Input Function Space for Each Region.}
\end{figure}

When constructing the discretized representation of the input function, a region-specific identifier is first assigned to each grid cell using binary encoding. Next, each grid point's spatial and temporal location is assigned the corresponding $x$ and $t$ values. Finally, initial and boundary conditions are separately added to the grid along the spatial and temporal dimensions. Although these conditions only directly apply to certain points (e.g., initial conditions at ${x_0}$ along the spatial dimension or boundary conditions at ${t_0}$ along the temporal dimension), they can propagate and extend to affect the entire grid. Therefore, the dimensionality of the input function space ${d_a}$ is determined by the number of sub-regions, as well as the number of initial and boundary conditions. It is worth noting that the encoding for each grid cell must incorporate the boundary and initial conditions from all regions in order to capture the influence of other regions on the cells in the current region. Additionally, since the spatial lengths of the regions vary, the grid sizes also differ across regions. However, the grid size in the temporal dimension $t$ remains consistent across all regions.

\section{Padding Method of the Grid Alignment Layer}\label{sup5}
The formula for the DFT is as follows:
\begin{equation}
X[k] = \sum\limits_{n = 0}^{N - 1} X [n] \cdot {e^{ - j\frac{{2\pi }}{N}kn}},\quad k = 0,1,2,...,N - 1
\label{eq1}
\end{equation}
where the discrete signal $X[n]$ has a length of $N$, and its frequency domain representation is denoted as $X[k]$, where $k$ is the frequency index. The term ${e^{ - j\frac{{2\pi }}{N}kn}}$ is the Fourier basis function, which is used to project the signal onto different frequency components, where $j$ is the imaginary unit.

Zero-padding neither introduces new frequency components in the frequency domain nor alters the existing ones; rather, it increases the number of frequency domain sampling points, thereby improving frequency resolution. By extending the signal length through zero-padding, the grid sizes of different regions can be aligned, ensuring that the frequency components from different regions are effectively coupled under the same Fourier basis functions.

Specifically, the grid dimension alignment layer ${{\mathcal A}_1}$ applies zero-padding to both ends of the Fourier Transform dimensions. Since the grid lengths are the same in certain dimensions (e.g., the time dimension $t$) but differ in others (e.g., the spatial dimension $x$), the zero-padding strategies for these two types of dimensions vary. The detailed method is presented in Eq.~\ref{eq2}.
\begin{equation}
\left\{ {\begin{array}{*{20}{c}}
{d_{head}^t = d_{end}^t = [{d_t} \times {r_t}]}\\
{\begin{array}{*{20}{c}}
{d_{head}^x = [d_x^{\max }*{r_x}] + [(d_x^{\max } - {d_{xe}})/2]}\\
{d_{end}^x = 2[d_x^{\max }*{r_x}] + d_x^{\max } - {d_{xe}} - d_{head}^x}
\end{array}}
\end{array}} \right.
\label{eq2}
\end{equation}
where ${r_x}$ and ${r_t}$ represent the padding ratios for the spatial and temporal dimensions, respectively; $[ \cdot ]$ is a rounding symbol. $d_{head}^t$ and $d_{end}^t$ represent the padding lengths at the beginning and end of the temporal dimension, while $d_{head}^t$ and $d_{end}^t$ represent the padding lengths at the start and end of the spatial dimension. ${d_{x,e}}$ is the original spatial dimension of region $e$, and $d_x^{\max } = \max \{ {d_{x1}},{d_{x2}}, \cdots ,{d_{xE}}\} $ is the maximum grid size in the spatial dimension.

\end{document}